\documentclass[letterpaper]{article} 
\usepackage{aaai2026}  
\usepackage{times}  
\usepackage{helvet}  
\usepackage{courier}  
\usepackage[hyphens]{url}  
\usepackage{graphicx} 
\usepackage{natbib}  
\usepackage{caption} 
\usepackage{algorithm}
\usepackage{algorithmic}

\usepackage{newfloat}
\usepackage{listings}
\DeclareCaptionStyle{ruled}{labelfont=normalfont,labelsep=colon,strut=off} 
\floatstyle{ruled}
\newfloat{listing}{tb}{lst}{}
\floatname{listing}{Listing}
\usepackage{booktabs}
\usepackage{amsmath}
\usepackage[table]{xcolor}
\usepackage{amsfonts}
\usepackage{multirow}
\newcommand{\ul}[1]{\underline{#1}}
\usepackage{makecell}

\usepackage{tocloft}

\usepackage{etoolbox}
\makeatletter
\AfterEndEnvironment{algorithm}{\let\@algcomment\relax}
\AtEndEnvironment{algorithm}{\kern2pt\hrule\relax\vskip3pt\@algcomment}
\let\@algcomment\relax
\newcommand\algcomment[1]{\def\@algcomment{\footnotesize#1}}
\makeatother

\title{Rethinking Visual Token Reduction in LVLMs Under Cross-Modal Misalignment}
\author{
    Rui Xu\textsuperscript{\rm 1}, Yunke Wang\textsuperscript{\rm 2}, Yong Luo\textsuperscript{\rm 1}\thanks{Corresponding authors: Yong Luo and Bo Du.}, Bo Du\textsuperscript{\rm 1}\footnotemark[1]
}
\affiliations{
    \textsuperscript{\rm 1}School of Computer Science, Wuhan University \\ \textsuperscript{\rm 2}School of Computer Science, The University of Sydney\\
    xurui7943@gmail.com, yunke.wang@sydney.edu.au, \{luoyong, dubo\}@whu.edu.cn
}
\usepackage{bibentry}

\begin{document}

\maketitle

\begin{abstract}
Large Vision-Language Models (LVLMs) encode visual inputs as dense sequences of patch-level tokens to capture fine-grained semantics. These visual tokens often outnumber their textual counterparts by a large margin, leading to substantial computational overhead and limiting the scalability of LVLMs in practice. Previous efforts have explored visual token reduction either prior to or within the large language models (LLMs). However, most in-LLM reduction approaches rely on text-conditioned interactions, implicitly assuming that textual tokens can reliably capture the importance of visual tokens. In this work, we revisit this assumption and reveal causal, semantic, and spatial forms of cross-modal misalignment. These misalignments undermine the effectiveness of text-guided visual token reduction. To address this, we introduce VisionDrop, a training-free, visual-only pruning framework that selects informative visual tokens based on intra-modal (visual-to-visual) attention, without relying on textual signals. To further suppress redundancy throughout the model hierarchy, we treat the visual encoder and the LLM as a unified system and design a progressive pruning pipeline. Our method performs dominant token selection and lightweight contextual merging at multiple stages, enabling fine-grained visual information to be retained even under aggressive token budgets. Extensive experiments across diverse benchmarks show that VisionDrop achieves consistent improvements over existing approaches, despite requiring no additional training or complex modifications. Notably, when integrated with LLaVA-NeXT-7B, VisionDrop achieves a $2.7\times$ reduction in inference latency and $6\times$ in FLOPs, while retaining 95.71\% of the original performance.
\end{abstract}

\begin{links}
    \link{Code}{https://github.com/Ruixxxx/VisionDrop}
\end{links}

\section{Introduction}
Large Vision-Language Models (LVLMs) have achieved remarkable progress across a wide range of multimodal tasks~\cite{neurips23/LLaVA,arxiv24/Qwen2-VL,arxiv25/Qwen2.5-VL,arxiv25/InternVL2.5,arxiv25/InternVL3}, such as visual question answering, image captioning, and visual reasoning. A key driver of this success is the use of dense patch-level tokenization, which encodes visual inputs as long sequences of tokens to capture rich fine-grained semantics. However, this dense representation comes at a significant computational cost. Compared to textual tokens, visual tokens often dominate the input sequence. It is common for a single image to be represented by hundreds or even thousands of tokens~\cite{cvpr24/LLaVA1.5}, leading to quadratic growth in attention computation~\cite{neurips17/attention}. This limits the scalability of LVLMs for high-resolution~\cite{liu2024llavanext} or real-time applications~\cite{arxiv24/Gemma,neurips25/vla-cache}.

\begin{figure}[!t]
  \centering
  \includegraphics[width=0.47\textwidth]{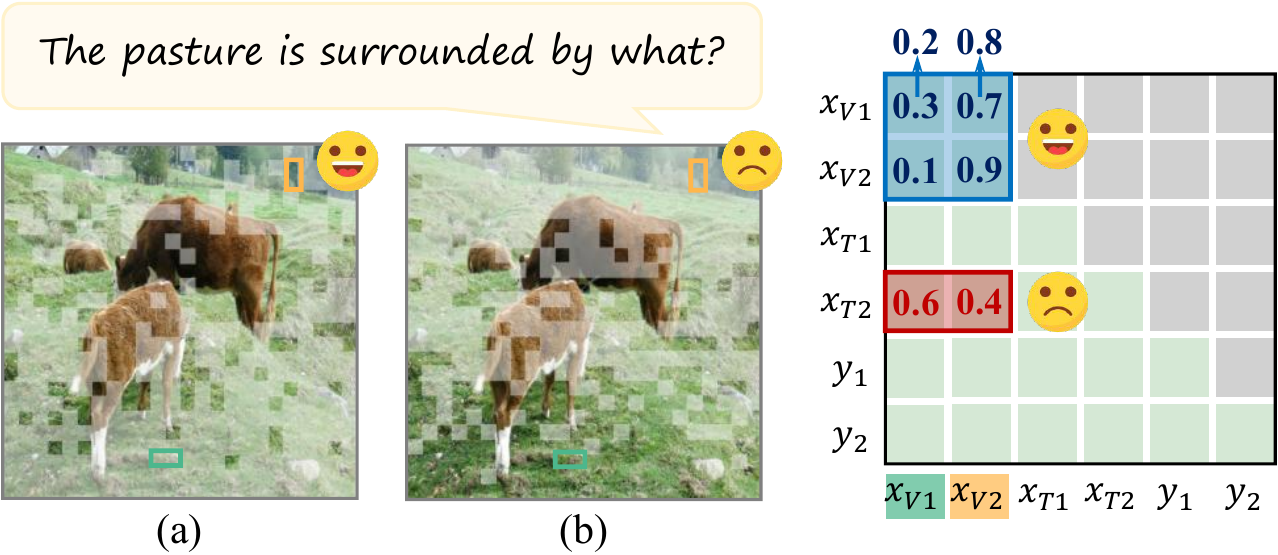}
  \caption{Comparison of visual token pruning strategies via attention maps from LLM decoding layers. Here, $x_{V1}, x_{V2}$ are visual tokens, $x_{T1}, x_{T2}$ are text instructions, and $y_1, y_2$ are autoregressively generated outputs. 
  (a) Our method identifies important visual tokens (e.g., the trees highlighted in orange) by leveraging image-to-image attention (blue box), which reflects intra-modal relevance and avoids interference from misaligned cross-modal signals. (b) In contrast, previous approaches rely on image-to-text attention (red box) to assess visual token importance, which can be overly sensitive to cross-modal noise, resulting in preservation of semantically redundant visual regions (e.g., the grass in green).
  }
  \label{fig:att_mask}
\end{figure}

To address this inefficiency, recent studies have explored reducing the number of visual tokens either before or within the layers of the large language models (LLMs)~\cite{cvpr25/VisionZip,icml25/SparseVLM,eccv24/FastV,cvpr25/PyramidDrop}. Pruning visual tokens before entering the LLM enables effective compression of redundant features at the representation level, but may risk discarding subtle or task-relevant details. In contrast, performing token reduction within the LLM allows uninformative tokens to be identified after visual information has been integrated into the language context, though it risks discarding entangled or contextually important features.
Additionally, a more fundamental limitation of many in-LLM pruning strategies lies in their reliance on text-conditioned scoring mechanisms, which estimate the relevance between visual tokens and textual inputs. They implicitly assume strong and persistent alignment between modalities throughout LLM layers. 

In this work, we revisit this critical assumption by asking: \textit{Are visual and textual representations truly well-aligned within the LLM layers?} Our analysis uncovers three types of cross-modal misalignment that challenge this assumption and limit the effectiveness of text-guided visual token pruning: (1) Causal misalignment arises from the autoregressive nature of LLMs, where the last text token tends to focus on nearby tokens in the input sequence, introducing locality bias in visual token scoring. (2) Semantic misalignment emerges as visual and textual tokens become deeply entangled within LLM layers, weakening the distinctiveness and interpretability of textual queries for assessing visual importance. (3) Spatial misalignment stems from the flattening of positional embeddings across modalities and the absence of spatial priors in textual inputs. While visual encoders already struggle to align spatial structures with textual semantics, this issue worsens in the LLM, where visual and textual tokens are fused into a single sequence. Text-guided pruning under such conditions may discard spatially important regions that are not explicitly emphasized by the text. 
While CLIP-style encoders~\cite{icml21/CLIP} and LVLM projection layers aim to align modalities at the input stage, such alignment degrades during multimodal token fusion in the LLM. As a result, pruning decisions based on text-guided signals may inadvertently discard visually salient or semantically important content.

To empirically validate this, we design a controlled study comparing visual-only and text-guided scoring strategies for visual token selection. Specifically, we replace image-to-text relevance scores with visual self-attention scores to assess the impact of modality alignment on pruning effectiveness across different compression ratios. Our findings reveal that, in general, visual-only scoring consistently outperforms text-guided scoring, especially under more aggressive pruning. This suggests that visual-textual alignment deteriorates as tokens propagate through the LLM, calling into question the reliability of text-guided reduction strategies.

Based on the analysis, we propose VisionDrop, a training-free framework for visual token reduction. Unlike prior approaches that rely on text-guided relevance, VisionDrop estimates token importance solely from visual self-attention, avoiding dependence on potentially misaligned textual cues. Our method performs stage-wise pruning across the full LVLM architecture, progressively reducing visual tokens in both the visual encoder and LLM layers. First, the dominant token selection identifies highly referenced visual tokens by the aforementioned visual-only importance score, ensuring that key semantic content is retained. Second, the lightweight contextual merging aggregates remaining tokens into contextual tokens by similarity, preserving complementary information. Together, these components are applied at multiple stages in the model, allowing VisionDrop to retain expressive visual representations under tight token budgets.

The contributions of our work are summarized as follows:

\begin{itemize}
    \item We empirically investigate the misalignment between visual and textual representations within LLM layers and offer insights to the visual token importance scoring.
    \item We propose VisionDrop, a training-free pruning framework that progressively prunes visual tokens across both the visual encoder and the LLM.
    \item We propose a visual-only scoring method for dominant token selection and apply the contextual token merging to preserve complementary information at each stage.
\end{itemize}

Extensive evaluations on diverse benchmarks show that VisionDrop outperforms prior state-of-the-art approaches. Under only 5.6\% token retention, it preserves 91.46\% and 92.06\% of the original LLaVA-1.5-7B~\cite{cvpr24/LLaVA1.5} and LLaVA-NeXT-7B~\cite{liu2024llavanext} performance, surpassing the best baseline by 0.96\% and 1.71\%, respectively.

\section{Related Works}

\subsection{Large Vision-Language Models}

The evolution of large vision-language models (LVLMs) has been driven by the impressive generalization and reasoning abilities of large language models (LLMs)~\cite{arxiv23/LLaMA,arxiv23/mistral7b,arxiv23/Qwen,arxiv25/Qwen2.5,arxiv24/InternLM2}, extending their success to the multimodal domain. Recent LVLMs integrate visual and textual modalities by projecting visual inputs into token sequences compatible with language models~\cite{neurips23/LLaVA,arxiv24/LLaVA-OneVision,arxiv24/Qwen2-VL,arxiv25/Qwen2.5-VL,arxiv25/InternVL2.5,arxiv25/InternVL3}, enabling them to leverage the full capacity of LLMs. However, the information density in visual data is often much lower than in textual data, resulting in an excessive number of visual tokens. For instance, LLaVA-1.5~\cite{cvpr24/LLaVA1.5} encodes a $336 \times 336$ image into 576 tokens, while LLaVA-NeXT~\cite{liu2024llavanext} expands this to 2,880 tokens for high-resolution input. This issue becomes even more critical in video understanding tasks. For example, LongVA~\cite{arxiv24/LongVA} transforms 2,000 frames into more than 200K visual tokens, while LongVILA~\cite{iclr25/LongVILA} supports inputs exceeding 6,000 frames and produces over a million visual tokens. The inflated visual token sequences place heavy demands on computation and memory, motivating the need for effective token reduction strategies to improve scalability and efficiency across tasks.

\begin{figure*}[!t]
  \centering
  \includegraphics[width=0.95\textwidth]{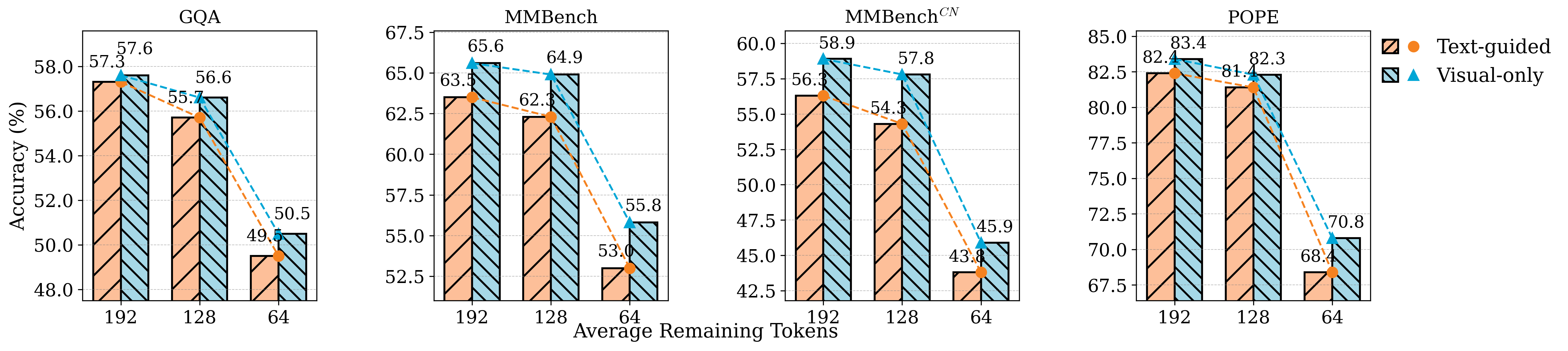}
  \caption{Accuracy comparisons between text-guided and visual-only scoring within the LLM across different average remaining token levels (192, 128, 64) on four benchmarks.
  }
  \label{fig:controlled_study}
\end{figure*}

\begin{figure}[!t]
  \centering
  \includegraphics[width=0.455\textwidth]{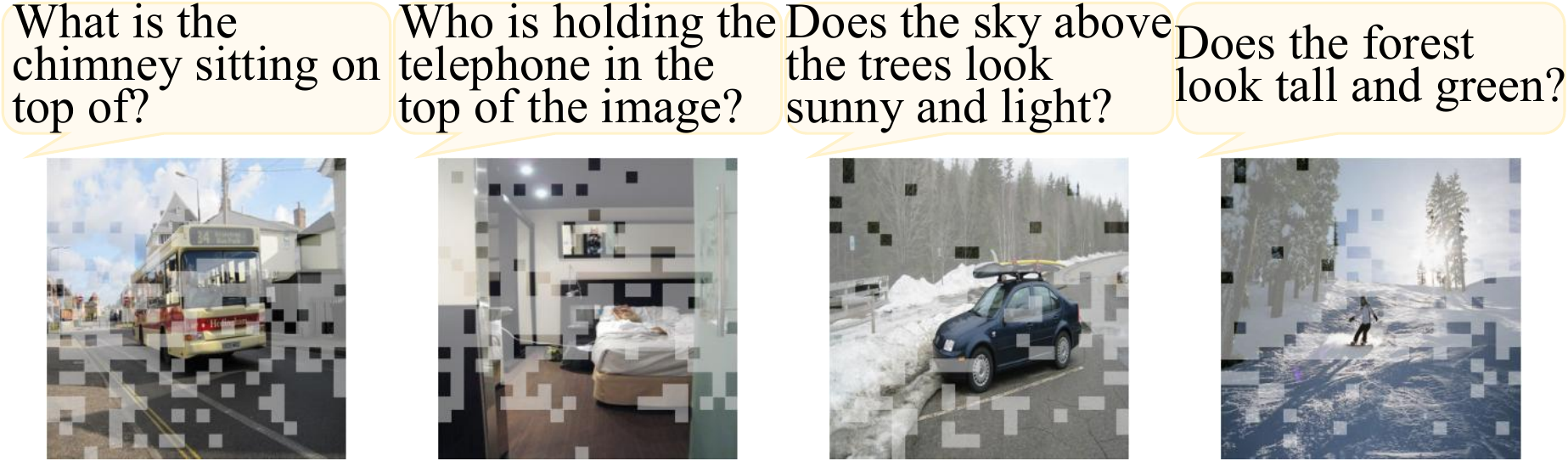}
  \caption{Visualization of text-guided visual token retention after shallow-layer pruning (specifically, the second layer) in the LLM. Pale translucent blocks indicate pruned tokens. Retained tokens consistently cluster at the bottom of the image, revealing a positional bias caused by the causal attention in autoregressive LLMs.}
  \label{fig:causal_issue}
\end{figure}

\begin{figure}[!t]
  \centering
  \includegraphics[width=0.455\textwidth]{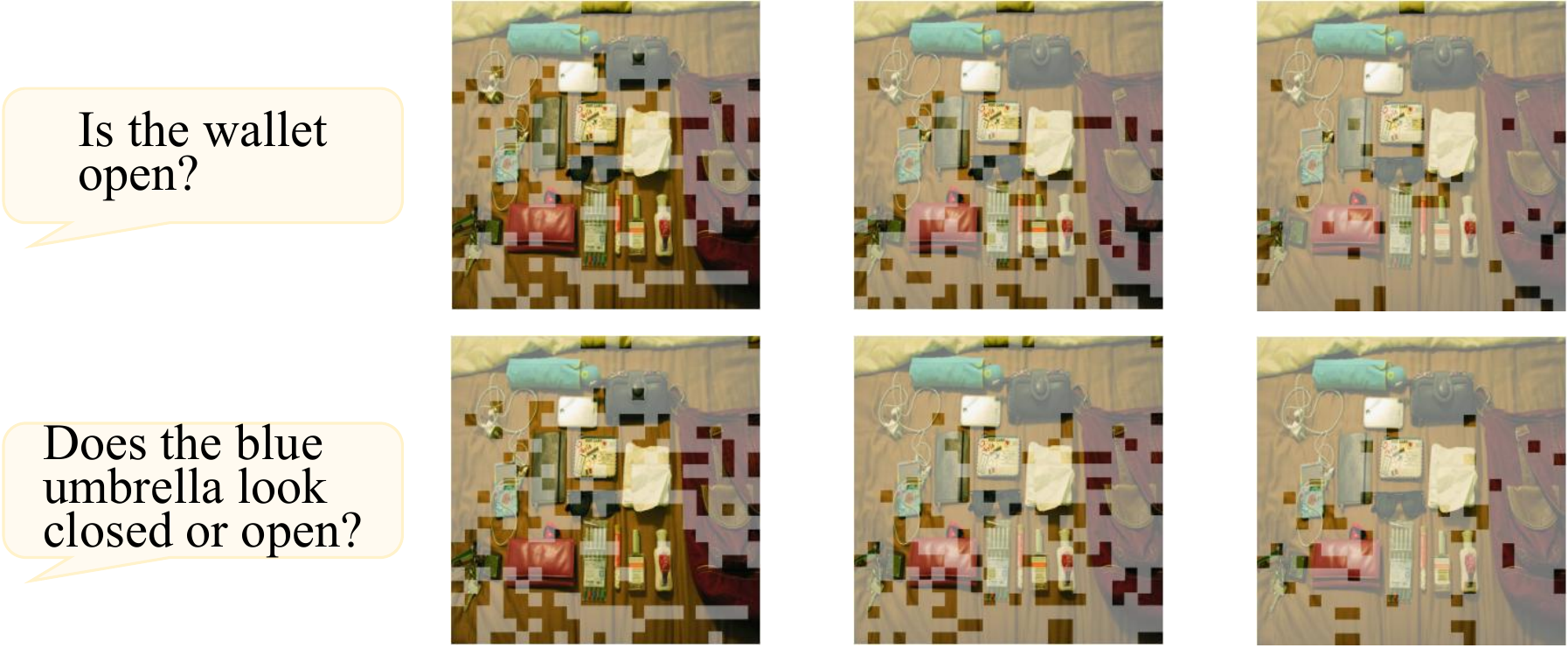}
  \caption{Visualization of text-guided visual token retention at the end of the first three LLM stages. While the wallet-related region is correctly preserved in response to the first question, the model fails to retain relevant tokens around the umbrella in the second case, reflecting the semantic misalignment caused by modal entanglement in LLM layers.}
  \label{fig:entangle_issue}
\end{figure}

\subsection{Visual Token Reduction in LVLMs}

To mitigate the computational burden in LVLMs, recent studies have explored reducing the number of visual tokens. Existing approaches to visual token reduction are primarily applied either in the visual encoder or within the LLM. They typically use either text-guided or text-agnostic scoring strategies to determine token importance. \textbf{Visual encoder pruning} approaches perform token reduction before tokens are fed into the LLM. Some approaches rely solely on visual information: FlowCut~\cite{arxiv25/FlowCut} leverages token-level information flow across layers to identify and prune redundant visual tokens in a progressive and structure-aware manner. VisionZip~\cite{cvpr25/VisionZip} and VisPruner~\cite{iccv25/VisPruner} select dominant tokens based on attention scores and compress the rest using similarity-based merging or pruning. VScan~\cite{arxiv25/VScan} applies global-local scanning. Other works utilize textual guidance at this stage. CDPruner~\cite{arxiv25/CDPruner} maximizes instruction-conditioned diversity of retained tokens. SparseVLM~\cite{icml25/SparseVLM} computes token importance via cross-modal attention with text guidance. Although visual encoder pruning benefits from more structured token semantics, it may risk losing fine-grained cues critical for multimodal reasoning. \textbf{LLM decoding pruning} is an alternative direction that performs token reduction within the LLM, where visual and textual tokens are jointly processed. Most of these approaches adopt text-guided token selection. FastV~\cite{eccv24/FastV} ranks tokens based on their attention from text during generation, and VScan also extends its pruning into LLM layers based on cross-modal attentions. PyramidDrop~\cite{cvpr25/PyramidDrop} selects visual tokens that are most relevant to the final instruction token, performing progressive pruning across LLM layers. However, these approaches rely on the assumption that visual and textual tokens remain semantically aligned within the LLM, a property that is not guaranteed due to modality entanglement introduced by joint self-attention~\cite{arxiv25/VmapL}.

\begin{figure*}[!t]
  \centering
  \includegraphics[width=0.9\textwidth]{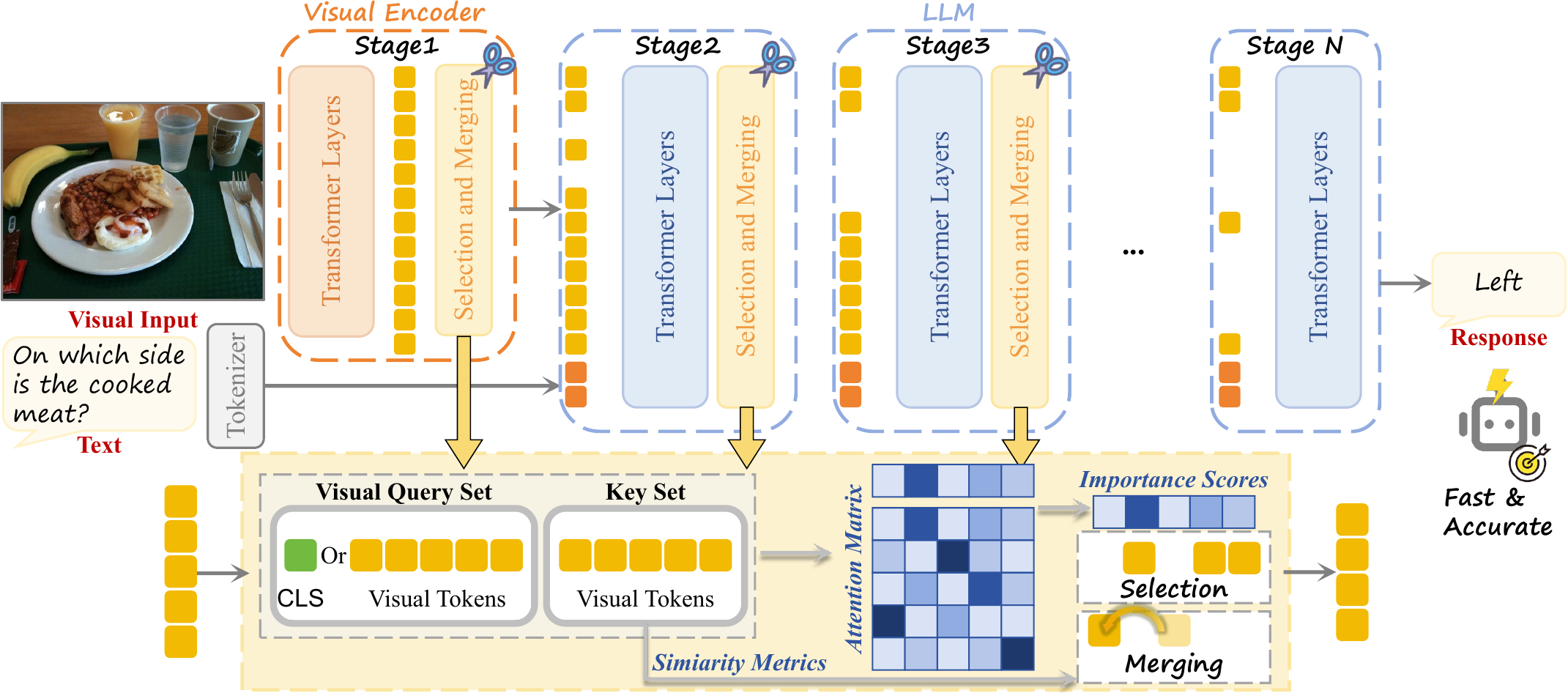}
  \caption{The visual encoder and LLM are partitioned into multiple pruning stages (e.g., Stage 1–N), where the number of visual tokens is progressively reduced. At the end of each stage, the importance of visual tokens is estimated via attention-based scores (values are denoted by color intensity, bottom), computed from self-attention among visual tokens (optionally utilizing the \texttt{[CLS]} as the query). Low-importance tokens are merged with each other using key-value similarity to form contextual representations, which are then propagated together with informative tokens to the next stage.}
  \label{fig:VisionDrop}
\end{figure*}

\section{Methodology}

\subsection{Rethinking Text-guided Visual Token Reduction}

To alleviate the computational overhead in LVLMs, recent studies have explored visual token reduction, either in the visual encoder~\cite{cvpr25/VisionZip,icml25/SparseVLM} or during the LLM decoding phase~\cite{eccv24/FastV,cvpr25/PyramidDrop}. In the latter case, mainstream approaches typically adopt text-guided strategies, using text relevance scores to determine which visual tokens to retain. However, this raises a critical question: \textit{Are visual and textual representations truly well-aligned within the LLM layers?} 
Although visual inputs are projected into the textual embedding space before entering the LLM, using text-guided strategies for visual token reduction introduces three forms of \textbf{cross-modal misalignment}, due to the modality-specific biases in how visual and textual information are processed.

\subsubsection{Causal.} 
In text-guided pruning, the last text instruction token is commonly used to assess the importance of visual tokens. However, due to the causal nature of autoregressive LLMs, this token tends to focus disproportionately on later-positioned visual tokens in the input sequence~\cite{arxiv25/VScan}. Unlike the \texttt{[CLS]} token or average-pooled visual features in bidirectional encoders, the instruction token does not serve as an explicit global aggregator. Figure~\ref{fig:causal_issue} illustrates this issue, where shallow-layer pruning results show that the retained visual tokens consistently cluster near the end of the input sequence, regardless of their semantic relevance.

\subsubsection{Semantic.} 
As tokens propagate through the LLM, visual and textual representations become increasingly entangled. The final instruction token evolves into a hybrid embedding that no longer maintains a clear semantic alignment with individual visual tokens, limiting its effectiveness in identifying visually important regions. This misalignment is evident in Figure~\ref{fig:entangle_issue}, where the retained visual tokens often fail to correspond to the semantics of the input question.

\subsubsection{Spatial.} 
Preserving spatially relevant visual information has long been a challenge in LVLMs~\cite{cvpr24/LION}. Even before entering the LLM, visual encoders (e.g., CLIP-style models~\cite{icml21/CLIP}) often struggle to align spatial structures with textual semantics, especially for tasks requiring fine-grained or spatial reasoning. This issue is further exacerbated within LLMs, where visual and textual tokens are flattened into a single sequence, and positional embeddings across modalities are merged, diluting the spatial priors of visual tokens. Moreover, textual inputs themselves lack spatial awareness and provide inherently incomplete representations of visual scenes. As a result, text-guided visual token pruning may amplify this spatial misalignment by discarding visually important regions that are not explicitly emphasized by the textual instruction.

These misalignments reveal that text-guided visual token pruning is fundamentally constrained by modality interaction biases. To empirically verify this, we conduct a controlled experiment by replacing text-guided scoring with visual-only scoring and observe the resulting performance changes. Specifically, we reproduce the PyramidDrop~\cite{cvpr25/PyramidDrop} framework. To ensure a fair comparison, we retain the same pruning configuration, including the pruning layers and token retention ratios, and replace the image-to-text relevance scores with image-to-image self-attention scores, thereby eliminating explicit textual guidance. We adopt LLaVA-v1.5-7B~\cite{cvpr24/LLaVA1.5} as the base LVLM and conduct extensive evaluations on 4 standard image understanding benchmarks, including GQA~\cite{cvpr19/GQA}, MMBench~\cite{eccv24/mmbench}, MMBench$^\text{CN}$~\cite{eccv24/mmbench}, and POPE~\cite{emnlp23/POPE}, to compare the effectiveness of the two scoring strategies. 
As shown in Figure~\ref{fig:controlled_study}, the image-only scoring strategy consistently outperforms the text-guided approach when retaining 192 visual tokens, and in general, the performance gap widens as the number of retained tokens decreases to 128 and 64. This performance gap under higher compression ratios indicates that visual and textual tokens are not well aligned within the LLM. Consequently, text-guided scoring, which implicitly assumes reliable cross-modal alignment, becomes increasingly unreliable when fewer visual tokens remain. In contrast, visual-only scoring exhibits greater robustness under aggressive pruning, highlighting its potential as a more stable and generalizable strategy for visual token reduction.

\subsection{Preliminaries}
LVLMs are commonly structured with three components: a vision encoder, a modality projector, and an LLM. Upon receiving the visual input, the vision encoder extracts patch-level features, which are subsequently mapped into a set of $n$ visual tokens $\mathbf{x}_V = \{x^i_V\}_{i=1}^{n}$ via the projector. These visual tokens are concatenated with the tokenized textual tokens $\mathbf{x}_T$ and fed into the LLM for autoregressive generation. At each timestep $t$, the model predicts the next token $y_t$ based on the conditional probability $p_\theta(y_t \mid \mathbf{x}_V, \mathbf{x}_T, \mathbf{y}{<t})$, where $\mathbf{y}{<t}$ denotes the previously generated tokens.

\subsection{VisionDrop}

Visual token reduction in LVLMs has been explored at two primary stages: either within the visual encoder~\cite{cvpr25/VisionZip,icml25/SparseVLM,iccv25/VisPruner} or during the LLM decoding phase~\cite{eccv24/FastV,cvpr25/PyramidDrop}. We observe that pruning at the encoder stage tends to be more stable, as visual tokens remain semantically coherent and well-structured. However, this may lead to the loss of fine-grained visual details that are crucial for downstream reasoning. In contrast, pruning in the LLM benefits from earlier visual-textual integration, where key visual cues may already be embedded in textual tokens, allowing more informed reduction. Yet as tokens propagate through LLM layers, visual information becomes increasingly entangled and diffused, making it difficult to reliably estimate token importance.

To address the cross-modal misalignment within LLMs and to leverage the complementary strengths of different pruning stages, we propose a training-free, visual-only pruning framework that performs progressive visual token reduction at both the encoder and LLM stages, balancing stability and expressiveness across the model hierarchy, as illustrated in Figure~\ref{fig:VisionDrop}.

\subsubsection{Progressive Dominant Token Selection.} 
We partition the LVLM architecture into a sequence of stages $\mathcal{S} = {s_0, s_1, \dots, s_N}$, including both the visual encoder and layers in the LLM decoder. At the end of each stage $s_n$, where $n = 1, 2, \dots, N$, we apply a stage-specific pruning ratio $\lambda_n$ to the current set of visual tokens $\mathbf{x}_V^{(s_n)}$, where $|\mathbf{x}_V^{(s_n)}| = \lambda_n \cdot |\mathbf{x}_V|$. Tokens are ranked based on visual-only attention-based importance scores, and the top-ranked subset is propagated to the next stage $s_{n+1}$. The selection is given by:

\begin{equation}
    \mathbf{x}_V^{(s_{n+1})} = \left\{ x_V^i \in \mathbf{x}_V^{(s_n)} \,\middle|\, S(i) \geq \tau_n \right\},
\end{equation}
where $S(i)$ denotes the visual importance score of token $x_V^i$, and $\tau_n$ is a threshold determined by $\lambda_n$.

To identify the most informative visual tokens, we compute importance scores by reusing the model's self-attention maps. Specifically, we measure how frequently each token is attended to by visual query tokens, averaged over all query positions. For clarity, we present the single-head attention formulation and omit the pruning stage notation.

Formally, let $\mathbf{x}_V^q \in \mathbb{R}^{L_1 \times D}$ denote the set of query visual tokens, and $\mathbf{x} \in \mathbb{R}^{L_2 \times D}$ be the full input sequence, including both visual and textual tokens. The attention query and key matrices are computed as:

\begin{equation}
\mathbf{Q} = \mathbf{x}_V^q \mathbf{W}_Q, \quad \mathbf{K} = \mathbf{x} \mathbf{W}_K,
\end{equation}
where $\mathbf{W}_Q, \mathbf{W}_K \in \mathbb{R}^{D \times D}$ are the projection matrices. The single-head attention matrix is obtained by:
\begin{equation}
\mathbf{A} = \text{Softmax} \left( \frac{\mathbf{Q} \mathbf{K}^\top}{\sqrt{D}} \right),
\end{equation}
where $\mathbf{A} \in \mathbb{R}^{L_1 \times L_2}$ contains attention scores from each query visual token to all tokens in the sequence.

To compute the final importance score $\mathbf{S} \in \mathbb{R}^{L_V}$ over the visual tokens, we first extract the attention weights $\mathbf{A}_{:, \mathcal{V}} \in \mathbb{R}^{L_1 \times L_V}$ corresponding to the visual key tokens from the full attention matrix $\mathbf{A} \in \mathbb{R}^{L_1 \times L_2}$. We then average across all visual queries:

\begin{equation}
\mathbf{S} = \frac{1}{L_1} \sum_{l=1}^{L_1} \mathbf{A}[l, \mathcal{V}],
\end{equation}

where $\mathcal{V}$ is the index set of visual tokens in the sequence.

\textit{Visual Query Selection.} In LLM, where no explicit \texttt{[CLS]} token exists, we compute visual-to-visual attention by isolating attention maps within the visual subspace. Specifically, we extract attention scores from image queries to full token sequence, where $\mathbf{Q}_V = \mathbf{x}_V \mathbf{W}_Q, \mathbf{K} = \mathbf{x} \mathbf{W}_K$. We then keep only $\mathbf{A}$ entries indexed by visual tokens.

In visual encoder, if a dedicated \texttt{[CLS]} token is present (e.g., CLIP~\cite{icml21/CLIP}), we use attention from it to each visual token as token importance, where $\mathbf{Q}_{\text{CLS}} = \mathbf{x}_{\text{CLS}} \mathbf{W}_Q$. If no such token exists (e.g., SigLIP~\cite{iccv23/SigLIP}), we follow a consistent strategy with LLM and average the attention each token receives from all visual tokens.

This unified formulation enables modality-consistent and architecture-adaptive token importance estimation.

\begin{table*}[t]
\centering
\small
\setlength{\tabcolsep}{6.4pt}
\begin{tabular}{l|cccccccccc}
\toprule
\textbf{Method}         & \textbf{GQA} & \textbf{MMB} & \textbf{MMB}$^{\text{CN}}$ & \textbf{MME} & \textbf{POPE} & \textbf{SQA} & \textbf{VQA}$^{\text{v2}}$ & \textbf{VQA}$^{\text{Text}}$ & \textbf{VizWiz} & \textbf{Avg.} \\
\hline
\rowcolor{gray!15}
\multicolumn{11}{c}{\textit{Upper Bound, 576 Tokens (100\%)}} \\
LLaVA-1.5-7B            & 61.92 & 66.31 & 58.63 & 1863 & 86.81 & 69.51 & 78.53 & 58.20 & 50.13 & 100\% \\
\hline
\rowcolor{gray!15}
\multicolumn{11}{c}{\textit{Retain 192 Tokens (\textdownarrow 66.7\%)}} \\
FastV (ECCV24)          & 52.62 & 57.74 & 48.43 & 1540 & 75.59 & 68.07 & 70.51 & 52.77 & 46.76 & 88.45\% \\
PDrop (CVPR25)          & 57.27 & 63.51 & 56.28 & 1778 & 82.40 & \bf{69.56} & 75.57 & 56.10 & 48.95 & 96.11\% \\
SparseVLM-v1.5 (ICML25) & 59.44 & \bf 65.41 & \bf 58.69 & \ul{1789} & 86.45 & 68.86 & \ul{77.01} & \ul{57.76} & 50.64 & \ul{98.64}\% \\
VisionZip (CVPR25)      & 59.25 & 64.46 & 57.29 & 1767 & 86.39 & 68.86 & 76.78 & 57.26 & \bf{51.55} & 98.11\% \\
VisPruner (ICCV25)      & \ul{59.52} & 64.91 & 57.62 & 1779 & \ul{86.57} & 68.52 & 76.96 & 57.42 & \ul{51.46} & 98.38\% \\
VisionDrop (Ours)       & \bf 59.99 & \ul{65.19} & \ul{58.41} & \bf 1801 & \bf 87.23 & \ul{69.06} & \bf 77.28 & \bf 57.81 & 50.01 & \textbf{98.76}\% \\
\hline
\rowcolor{gray!15}
\multicolumn{11}{c}{\textit{Retain 128 Tokens (\textdownarrow 77.8\%)}} \\
FastV (ECCV24)          & 51.83 & 57.29 & 49.05 & 1502 & 75.55 & 68.37 & 66.61 & 51.09 & 49.41 & 87.88\% \\
PDrop (CVPR25)          & 55.70 & 62.28 & 54.32 & 1656 & 81.43 & \bf 69.36 & 74.21 & 55.32 & 49.01 & 94.04\% \\
SparseVLM-v1.5 (ICML25) & 58.43 & \bf 65.25 & \bf 58.86 & 1750 & \bf 86.37 & 68.57 & \ul{76.21} & 56.61 & 50.27 & \ul{97.76}\% \\
VisionZip (CVPR25)      & 57.62 & 63.40 & 56.67 & \ul{1768} & 84.69 & 68.82 & 75.60 & 56.81 & \ul{52.02} & 97.16\% \\
VisPruner (ICCV25)      & \ul{58.48} & 63.57 & 56.67 & 1743 & 85.67 & \ul{69.01} & 75.74 & \ul{57.00} & \bf 52.60 & 97.53\% \\
VisionDrop (Ours)       & \bf 58.61 & \ul{64.52} & \ul{57.06} & \bf 1777 & \ul{85.92} & 68.52 & \bf 76.24 & \bf 57.63 & 51.06 & \textbf{97.80}\% \\
\hline
\rowcolor{gray!15}
\multicolumn{11}{c}{\textit{Retain 64 Tokens (\textdownarrow 88.9\%)}} \\
FastV (ECCV24)          & 48.70 & 52.30 & 43.27 & 1409 & 67.21 & 69.36 & 61.08 & 49.69 & 50.55 & 83.13\% \\
PDrop (CVPR25)          & 49.51 & 52.97 & 43.83 & 1429 & 68.42 & \bf 69.96 & 63.04 & 50.80 & 50.02 & 84.23\% \\
SparseVLM-v1.5 (ICML25) & 53.77 & 61.27 & 52.13 & 1591 & 80.86 & \ul{69.61} & 70.24 & 53.46 & 50.35 & 92.06\% \\
VisionZip (CVPR25)      & 55.16 & 62.61 & 54.20 & \ul{1687} & 80.45 & 69.01 & 72.42 & 55.48 & \ul{52.94} & 94.62\% \\
VisPruner (ICCV25)      & \ul{55.59} & \ul{62.56} & \ul{54.48} & 1679 & \ul{81.29} & 68.82 & \ul{72.64} & \bf 55.83 & \bf 53.00 & \ul{94.89}\% \\
VisionDrop (Ours)       & \bf 55.89 & \bf 62.95 & \bf 55.10 & \bf 1698 & \bf 81.58 & 69.31 & \bf 73.16 & \ul{55.59} & 52.28 & \textbf{95.22}\% \\
\hline
\rowcolor{gray!15}
\multicolumn{11}{c}{\textit{Retain 32 Tokens (\textdownarrow 94.4\%)}} \\
VisionZip (CVPR25)      & 51.80 & 58.02 & 49.78 & 1592 & 75.11 & 68.72 & 67.12 & 53.17 & \bf 52.91 & 89.92\% \\
VisPruner (ICCV25)      & 51.98 & 59.02 & 50.78 & \bf 1593 & 76.44 & 69.26 & 67.41 & 53.47 & 52.16 & 90.50\% \\
VisionDrop (Ours)       & \bf 52.79 & \bf 60.31 & \bf 52.91 & 1572 & \bf 77.19 & \bf 69.41 & \bf 68.55 & \bf 53.56 & 52.26 & \textbf{91.46}\% \\
\bottomrule
\end{tabular}
\caption{Performance comparison on LLaVA-1.5-7B~\cite{cvpr24/LLaVA1.5} under different token retention rates. The best results are marked in \textbf{bold}, and the second-best are \ul{underlined}.}
\label{tab:image_main_results}
\end{table*}

\subsubsection{Stage-wise Contextual Token Merging.} 
While dominant tokens capture the primary visual content, discarding the remaining ones may result in missing subtle or auxiliary visual cues. To address this, we introduce a lightweight merging step at each pruning stage to preserve such complementary information. While similar ideas have been proposed in prior work~\cite{cvpr25/VisionZip}, our framework uniquely generalizes this operation across both the visual encoder and LLM decoder stages in a progressive fashion. 

Specifically, we reuse the key embeddings from the attention module, which encode semantic content of tokens, to measure pairwise similarity. In the LLM, we explicitly extract the image-token portion of the key vectors before similarity computation, ensuring modality-pure merging. At the end of each stage, the non-dominant tokens are divided into candidate and reference sets, and each candidate token is matched to its most similar reference token based on a dot-product similarity. The matched tokens are then fused to produce enriched contextual tokens. This process ensures that visually redundant tokens are merged rather than dropped, preserving fine-grained details while keeping the token count under control.

\begin{table*}[t]
\centering
\small
\setlength{\tabcolsep}{7.2pt}
\begin{tabular}{l|cccccccccc}
\toprule
\textbf{Method}      & \textbf{GQA} & \textbf{MMB} & \textbf{MMB}$^{\text{CN}}$ & \textbf{MME} & \textbf{POPE} & \textbf{SQA} & \textbf{VQA}$^{\text{v2}}$ & \textbf{VQA}$^{\text{Text}}$ & \textbf{VizWiz} & \textbf{Avg.} \\
\hline
\rowcolor{gray!15}
\multicolumn{11}{c}{\textit{Upper Bound, 2880 Tokens (100\%)}} \\
LLaVA-NeXT-7B        & 64.27 & 69.39 & 62.00 & 1854 & 87.54 & 70.30 & 81.83 & 61.40 & 57.63 & 100\% \\
\hline
\rowcolor{gray!15}
\multicolumn{11}{c}{\textit{Retain 640 Tokens (\textdownarrow 77.8\%)}} \\
FastV (ECCV24)       & 56.26 & 64.97 & 59.25 & 1646 & 83.98 & 68.27 & 73.90 & 54.80 & 55.00 & 92.62\% \\
PDrop (CVPR25)       & \bf 61.80 & \bf 68.11 & \bf 61.10 & 1782 & 86.44 & \bf 69.36 & \ul{79.71} & 59.44 & 55.44 & \ul{97.42}\% \\
VisionZip (CVPR25)   & 61.31 & 66.14 & \ul{60.43} & \ul{1805} & \ul{87.38} & 67.82 & 79.13 & \ul{60.18} & \bf 57.32 & 97.33\% \\
VisionDrop (Ours)    & \ul{61.70} & \ul{66.98} & \ul{60.43} & \bf 1807 & \bf 87.85 & \ul{68.32} & \bf 79.76 & \bf 60.45 & \ul{56.78} & \textbf{97.72}\% \\
\hline
\rowcolor{gray!15}
\multicolumn{11}{c}{\textit{Retain 320 Tokens (\textdownarrow 88.9\%)}} \\
FastV (ECCV24)       & 53.49 & 61.94 & 54.99 & 1495 & 76.16 & 66.83 & 68.24 & 51.40 & 52.07 & 86.82\% \\
PDrop (CVPR25)       & 57.08 & \ul{64.46} & 56.67 & 1672 & 79.85 & \bf 69.91 & 74.81 & 54.57 & 53.66 & 91.93\% \\
VisionZip (CVPR25)   & \ul{58.92} & 63.40 & \ul{57.40} & \ul{1712} & \ul{84.63} & 67.67 & \ul{76.32} & \ul{58.86} & \ul{56.64} & \ul{94.26}\% \\
VisionDrop (Ours)    & \bf 59.95 & \bf 64.91 & \bf 58.13 & \bf 1789 & \bf 85.21 & \ul{67.97} & \bf 77.67 & \bf 59.24 & \bf 56.97 & \textbf{95.71}\% \\
\hline
\rowcolor{gray!15}
\multicolumn{11}{c}{\textit{Retain 160 Tokens (\textdownarrow 94.4\%)}} \\
VisionZip (CVPR25)   & 55.55 & 60.09 & 54.71 & 1626 & 79.63 & \bf 68.12 & 71.72 & 56.24 & 55.95 & 90.35\% \\
VisionDrop (Ours)    & \bf 57.02 & \bf 62.00 & \bf 55.61 & \bf 1646 & \bf 81.88 & 67.53 & \bf 73.99 & \bf 57.52 & \bf 56.65 & \textbf{92.06}\% \\
\bottomrule
\end{tabular}
\caption{Performance comparison on LLaVA-NeXT-7B~\cite{liu2024llavanext} under different token retention rates.}
\label{tab:image_main_results_next}
\end{table*}

\section{Experiments}

\subsection{Experimental Setups}

\subsubsection{Models and Baselines.} 
We evaluate the proposed VisionDrop on the widely adopted LLaVA-1.5-7B~\cite{liu2024llavanext} and LLaVA-NeXT-7B~\cite{liu2024llavanext} for image understanding, as well as Video-LLaVA-7B~\cite{emnlp24/videollava} for video understanding. We compare our method with several state-of-the-art visual token reduction baselines, including FastV~\cite{eccv24/FastV}, PyramidDrop~\cite{cvpr25/PyramidDrop}, SparseVLM~\cite{icml25/SparseVLM}, VisionZip~\cite{cvpr25/VisionZip}, and VisPruner~\cite{iccv25/VisPruner}.

\subsubsection{Benchmarks.} 
We evaluate VisionDrop on 9 standard image understanding benchmarks. These include general visual question answering datasets such as GQA~\cite{cvpr19/GQA}, VQAv2~\cite{cvpr17/VQAv2}, and TextVQA~\cite{cvpr19/textvqa}; robustness and fairness benchmarks including MMBench~\cite{eccv24/mmbench}, MMBench-CN~\cite{eccv24/mmbench}, MME~\cite{arxiv24/MME} and POPE~\cite{emnlp23/POPE}; as well as specialized datasets like VizWiz~\cite{cvpr18/Vizwiz}, which features low-quality images, and ScienceQA (SQA)~\cite{neurips22/sqa}, which focuses on scientific reasoning. We also evaluate using 3 video understanding benchmarks, TGIF~\cite{cvpr17/tgif}, MSVD~\cite{acl11/msvd}, and MSRVTT~\cite{cvpr16/msrvtt}.

\subsubsection{Implementation Details.}
We adopt the standard inference settings provided in the official implementations of each LVLM. For progressive visual token pruning, we divide the model into five stages: the first stage concludes at the output of the visual encoder, while the remaining four end at decoding layers $l=8$, $16$, $24$, and the final decoding layer of the LLM. The number of visual tokens is gradually reduced across these stages. In the first stage, all original visual tokens are retained. 
For each reduction configuration, the retention number in the second stage is set to $1.5\times$ the final count for image understanding and $3\times$ for video understanding, and tokens are pruned proportionally in subsequent stages to meet the overall budget.

\subsection{Results and Analyses}

\subsubsection{Results on LLaVA-1.5.} 
As shown in Table~\ref{tab:image_main_results}, VisionDrop consistently outperforms all competing approaches across different token retention levels. Even with only 32 visual tokens (a 94.4\% reduction), it preserves 91.46\% of the full-token performance, demonstrating strong efficiency with minimal degradation. While SparseVLM~\cite{icml25/SparseVLM} performs well at higher token counts but degrades under extreme compression, VisPruner~\cite{iccv25/VisPruner} and VisionZip~\cite{cvpr25/VisionZip} show better robustness at low budgets, performing particularly well on VizWiz~\cite{cvpr18/Vizwiz}, likely due to their early-stage pruning in low-quality images. Remarkably, VisionDrop maintains the best performance, especially under tighter token constraints.

\subsubsection{Results on LLaVA-NeXT.} 
LLaVA-NeXT~\cite{liu2024llavanext} enhances LLaVA-1.5 by supporting higher-resolution visual inputs, yielding more visual tokens and redundancy, making it a compelling testbed for evaluating token pruning strategies. As shown in Table~\ref{tab:image_main_results_next}, VisionDrop consistently outperforms all baselines across different token retention rates, demonstrating its strong ability to retain critical visual information. Notably, its advantage becomes more pronounced as the number of retained tokens decreases. At 160 tokens (a 94.4\% reduction), VisionDrop achieves an average performance of 92.06\%, which is 1.71\% higher than the second-best approach.

\begin{table}[t]
\centering
\small
\setlength{\tabcolsep}{1.8pt}
\begin{tabular}{l|cc|cc|cc|cc}
\toprule
\multirow{2}{*}{\textbf{Method}} 
& \multicolumn{2}{c|}{\textbf{TGIF}} 
& \multicolumn{2}{c|}{\textbf{MSVD}} 
& \multicolumn{2}{c|}{\textbf{MSRVTT}} 
& \multicolumn{2}{c}{\textbf{Avg.}} \\ 
\cline{2-9}
& Acc. & Score & Acc. & Score & Acc. & Score & Acc. & Score \\ 
\hline
Video-LLaVA   & 18.9 & 2.52 & 71.6 & 3.94 & 56.8 & 3.48 & 49.1 & 3.31 \\ 
FastV         & 23.2 & 2.48 & 46.6 & 3.21 & 45.0 & 3.14 & 38.3 & 2.94 \\ 
VisionZip     & 15.6 & 2.40 & 70.2 & 3.94 & 54.0 & 3.41 & 46.6 & 3.25 \\ 
VisionDrop    & 18.8 & 2.48 & 68.4 & 3.87 & 54.6 & 3.42 & \textbf{47.3} & \textbf{3.26} \\ 
\bottomrule
\end{tabular}
\caption{Performance comparison on Video-LLaVA-7B~\cite{emnlp24/videollava} under a 256 token retention.}
\label{tab:video_main_result}
\end{table}

\subsubsection{Results on Video-LLaVA.} 
We further evaluate VisionDrop on video understanding tasks. Following prior works~\cite{eccv24/FastV}, we use the first 1,000 samples from each dataset for evaluation due to API usage limits. As shown in Table~\ref{tab:video_main_result}, VisionDrop achieves the highest average performance among all compared approaches, obtaining an accuracy of 47.3\% and a score of 3.26, while retaining only 12.5\% of the visual tokens.

\begin{figure}[!t]
  \centering
  \includegraphics[width=0.47\textwidth]{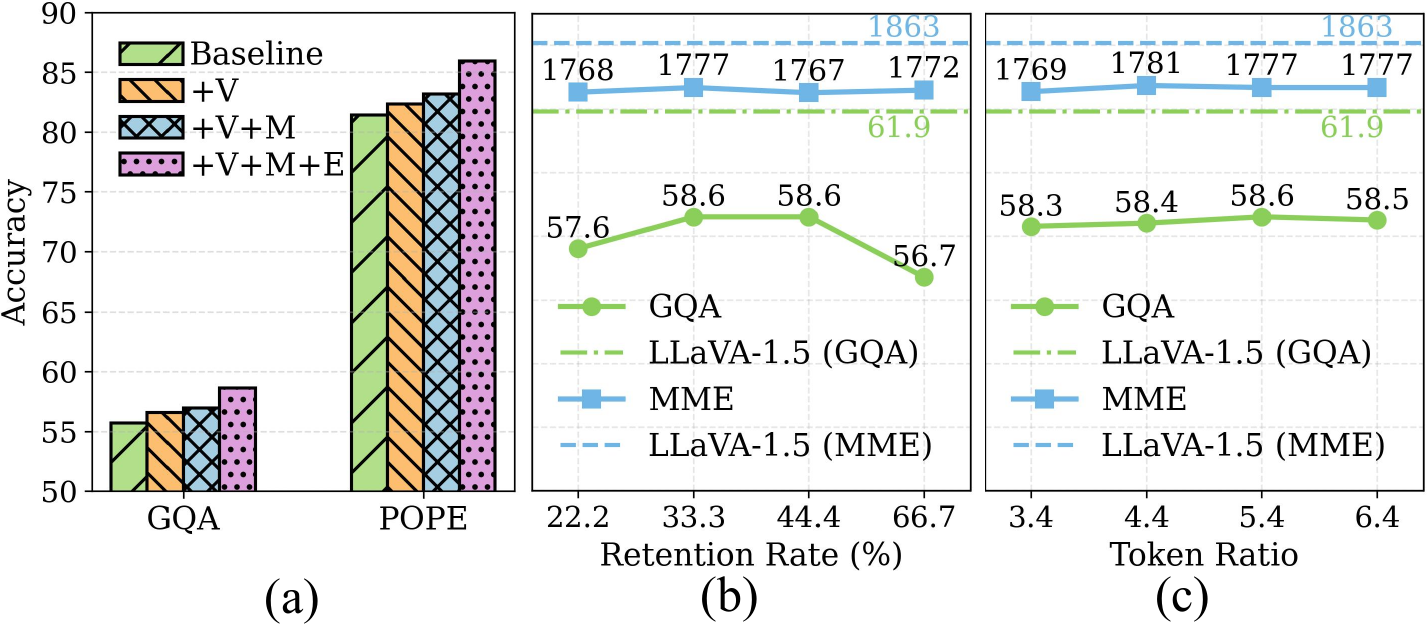}
  \caption{Ablation studies under 22.2\% token retention. (a) Component-wise results. V: visual-only; M: Merging; E: Encoder pruning. (b) Token retention rates applied to the visual encoder. (c) Dominant-to-contextual token ratios.
  }
  \label{fig:ablation_study}
\end{figure}

\begin{table}[t]
\centering
\small
\setlength{\tabcolsep}{1.2pt}
\begin{tabular}{l|c|ccc}
\toprule
\textbf{Method} & \textbf{\# Token} & \textbf{FLOPs (T)} & \textbf{Memory (GB)} & \textbf{Time (ms)} \\
\hline
LLaVA-1.5     & 576  & 9.06 & 14.52 & 237 \\
VisionDrop    & 64   & 2.11 & 14.35 & 117 ($\times$2.0) \\
\hline
LLaVA-NeXT    & 2880 & 46.25 & 16.56 & 593 \\
VisionDrop    & 320  & 7.70 & 14.63 & 216 ($\times$2.7) \\
\bottomrule
\end{tabular}
\caption{Efficiency analysis of VisionDrop on LLaVA-1.5-7B and LLaVA-NeXT-7B.}
\label{tab:efficiency}
\end{table}

\subsubsection{Ablation Study.} 
We conduct ablation studies under a 22.2\% token retention setting. Figure~\ref{fig:ablation_study}(a) presents the results of different component combinations of our method. The results demonstrate that all individual modules contribute positively to performance. Specifically, incorporating visual-only scoring, by removing dependence on text queries within the LLM, yields a notable improvement. The inclusion of stage-wise token merging further boosts performance by retaining fine-grained visual context that would otherwise be lost. 
Figure~\ref{fig:ablation_study}(b) examines the impact of different token retention rates in the visual encoder. It is observed that a moderate retention rate of 33.3\% achieves the best performance.  Figure~\ref{fig:ablation_study}(c) further explores the effect of varying the dominant-to-contextual token ratios. The model maintains stable performance across ratios, with slight improvements when dominant tokens are emphasized.

\subsubsection{Efficiency Analysis.} 
To evaluate the efficiency of VisionDrop, we conduct a comparative analysis of FLOPs, memory usage, and inference latency on LLaVA-1.5 and LLaVA-NeXT. All experiments are conducted on a single 24GB NVIDIA GeForce RTX 3090 GPU. As shown in Table~\ref{tab:efficiency}, by retaining only 64 and 320 tokens for LLaVA-1.5 and LLaVA-NeXT, respectively, and preserving over 95\% of the original performance, VisionDrop achieves a $2.0\times$ and $2.7\times$ speedup, along with a $4.3\times$ and $6.0\times$ reduction in FLOPs.

\section{Conclusion}
In this work, we introduce VisionDrop, a simple yet effective training-free, visual-only pruning framework designed to mitigate modality misalignment and reduce computational cost in LVLMs. By treating the visual encoder and the LLM as a unified processing pipeline, VisionDrop performs progressive dominant token selection and contextual merging without relying on text-conditioned signals. This allows the model to preserve critical visual cues under strict token budgets. Moreover, VisionDrop’s reliance on purely visual signals makes it especially advantageous in domains where language cues are sparse or weakly aligned with visual content, such as high-resolution medical or remote sensing images.

\section{Acknowledgments}
This research was supported in part by the National Natural Science Foundation of China (Grant Nos. U23A20318, 62276195, and 62225113).

\bibliography{aaai2026_visiondrop}

@inproceedings{neurips23/LLaVA,
  author       = {Haotian Liu and
                  Chunyuan Li and
                  Qingyang Wu and
                  Yong Jae Lee},
  title        = {Visual Instruction Tuning},
  booktitle    = {NeurIPS},
  year         = {2023}
}

@article{arxiv24/LLaVA-OneVision,
  	title={LLaVA-OneVision: Easy Visual Task Transfer},
  	author={Li, Bo and Zhang, Yuanhan and Guo, Dong and Zhang, Renrui and Li, Feng and Zhang, Hao and Zhang, Kaichen and Li, Yanwei and Liu, Ziwei and Li, Chunyuan},
  	journal={arXiv preprint arXiv:2408.03326},
  	year={2024}
}

@article{arxiv24/Qwen2-VL,
  title={Qwen2-VL: Enhancing Vision-Language Model's Perception of the World at Any Resolution},
  author={Wang, Peng and Bai, Shuai and Tan, Sinan and Wang, Shijie and Fan, Zhihao and Bai, Jinze and others},
  journal={arXiv preprint arXiv:2409.12191},
  year={2024}
}

@article{arxiv25/Qwen2.5-VL,
  title={Qwen2.5-VL Technical Report},
  author={Bai, Shuai and Chen, Keqin and Liu, Xuejing and Wang, Jialin and Ge, Wenbin and Song, Sibo and others},
  journal={arXiv preprint arXiv:2502.13923},
  year={2025}
}

@article{arxiv25/InternVL2.5,
      title={Expanding Performance Boundaries of Open-Source Multimodal Models with Model, Data, and Test-Time Scaling}, 
      author={Zhe Chen and Weiyun Wang and Yue Cao and Yangzhou Liu and Zhangwei Gao and Erfei Cui and others},
      journal={arXiv preprint arXiv:2412.05271},
      year={2025}
}

@article{arxiv25/InternVL3,
      title={InternVL3: Exploring Advanced Training and Test-Time Recipes for Open-Source Multimodal Models}, 
      author={Jinguo Zhu and Weiyun Wang and Zhe Chen and Zhaoyang Liu and Shenglong Ye and Lixin Gu and others},
      journal={arXiv preprint arXiv:2504.10479},
      year={2025}
}

@article{arxiv23/LLaMA,
      title={LLaMA: Open and Efficient Foundation Language Models}, 
      author={Hugo Touvron and Thibaut Lavril and Gautier Izacard and Xavier Martinet and Marie-Anne Lachaux and Timothée Lacroix and others},
      journal={arXiv preprint arXiv:2302.13971},
      year={2023}
}

@article{arxiv23/mistral7b,
      title={Mistral 7B}, 
      author={Albert Q. Jiang and Alexandre Sablayrolles and Arthur Mensch and Chris Bamford and Devendra Singh Chaplot and Diego de las Casas and others},
      journal={arXiv preprint arXiv:2310.06825},
      year={2023} 
}

@article{arxiv23/Qwen,
      title={Qwen Technical Report}, 
      author={Jinze Bai and Shuai Bai and Yunfei Chu and Zeyu Cui and Kai Dang and Xiaodong Deng and others},
      journal={arXiv preprint arXiv:2309.16609},      
      year={2023} 
}

@article{arxiv25/Qwen2.5,
      title={Qwen2.5 Technical Report}, 
      author={Qwen and : and An Yang and Baosong Yang and Beichen Zhang and Binyuan Hui and Bo Zheng and Bowen Yu and others},
      journal={arXiv preprint arXiv:2412.15115},
      year={2025}
}

@article{arxiv24/InternLM2,
      title={InternLM2 Technical Report}, 
      author={Zheng Cai and Maosong Cao and Haojiong Chen and Kai Chen and Keyu Chen and Xin Chen and others},
      journal={arXiv preprint arXiv:2403.17297},
      year={2024}
}

@inproceedings{cvpr24/LLaVA1.5,
  author       = {Haotian Liu and
                  Chunyuan Li and
                  Yuheng Li and
                  Yong Jae Lee},
  title        = {Improved Baselines with Visual Instruction Tuning},
  booktitle    = {CVPR},
  pages        = {26286--26296},
  year         = {2024}
}

@misc{liu2024llavanext,
    title={LLaVA-NeXT: Improved reasoning, OCR, and world knowledge},
    url={https://llava-vl.github.io/blog/2024-01-30-llava-next/},
    author={Liu, Haotian and Li, Chunyuan and Li, Yuheng and Li, Bo and Zhang, Yuanhan and Shen, Sheng and Lee, Yong Jae},
    month={January},
    year={2024}
}

@article{arxiv24/LongVA,
      title={Long Context Transfer from Language to Vision}, 
      author={Peiyuan Zhang and Kaichen Zhang and Bo Li and Guangtao Zeng and Jingkang Yang and Yuanhan Zhang and others},
      journal={arXiv preprint arXiv:2406.16852},
      year={2024}
}

@inproceedings{iclr25/LongVILA,
    title={Long{VILA}: Scaling Long-Context Visual Language Models for Long Videos},
    author={Yukang Chen and Fuzhao Xue and Dacheng Li and Qinghao Hu and Ligeng Zhu and Xiuyu Li and others},
    booktitle={ICLR},
    year={2025}
}

@article{arxiv25/FlowCut,
      title={FlowCut: Rethinking Redundancy via Information Flow for Efficient Vision-Language Models}, 
      author={Jintao Tong and Wenwei Jin and Pengda Qin and Anqi Li and Yixiong Zou and Yuhong Li and Yuhua Li and Ruixuan Li},
      journal={arXiv preprint arXiv:2505.19536},
      year={2025},
}

@inproceedings{cvpr25/VisionZip,
  title={VisionZip: Longer is Better but Not Necessary in Vision Language Models},
  author={Yang, Senqiao and Chen, Yukang and Tian, Zhuotao and Wang, Chengyao and Li, Jingyao and Yu, Bei and Jia, Jiaya},
  booktitle={CVPR},
  year={2025}
}

@article{arxiv25/VScan,
      title={VScan: Rethinking Visual Token Reduction for Efficient Large Vision-Language Models}, 
      author={Ce Zhang and Kaixin Ma and Tianqing Fang and Wenhao Yu and Hongming Zhang and Zhisong Zhang and Yaqi Xie and Katia Sycara and Haitao Mi and Dong Yu},
      journal={arXiv preprint arXiv:2505.22654},
      year={2025}
}

@article{arxiv25/CDPruner,
      title={Beyond Attention or Similarity: Maximizing Conditional Diversity for Token Pruning in MLLMs}, 
      author={Qizhe Zhang and Mengzhen Liu and Lichen Li and Ming Lu and Yuan Zhang and Junwen Pan and Qi She and Shanghang Zhang},
      journal={arXiv preprint arXiv:2506.10967},
      year={2025}
}

@inproceedings{icml25/SparseVLM,
  title={SparseVLM: Visual Token Sparsification for Efficient Vision-Language Model Inference},
  author={Zhang, Yuan and Fan, Chun-Kai and Ma, Junpeng and Zheng, Wenzhao and Huang, Tao and Cheng, Kuan and Gudovskiy, Denis and Okuno, Tomoyuki and Nakata, Yohei and Keutzer, Kurt and others},
  booktitle={ICML},
  year={2025}
}

@inproceedings{eccv24/FastV,
      title={An Image is Worth 1/2 Tokens After Layer 2: Plug-and-Play Inference Acceleration for Large Vision-Language Models}, 
      author={Liang Chen and Haozhe Zhao and Tianyu Liu and Shuai Bai and Junyang Lin and Chang Zhou and Baobao Chang},
      year={2024},
      booktitle={ECCV}
}

@inproceedings{cvpr25/PyramidDrop,
      title={PyramidDrop: Accelerating Your Large Vision-Language Models via Pyramid Visual Redundancy Reduction}, 
      author={Long Xing and Qidong Huang and Xiaoyi Dong and Jiajie Lu and Pan Zhang and Yuhang Zang and Yuhang Cao and Conghui He and Jiaqi Wang and Feng Wu and Dahua Lin},
      booktitle={CVPR},      
      year={2025} 
}

@inproceedings{icml21/CLIP,
  author       = {Alec Radford and
                  Jong Wook Kim and
                  Chris Hallacy and
                  Aditya Ramesh and
                  Gabriel Goh and
                  Sandhini Agarwal and
                  Girish Sastry and
                  Amanda Askell and
                  Pamela Mishkin and
                  Jack Clark and
                  Gretchen Krueger and
                  Ilya Sutskever},
  title        = {Learning Transferable Visual Models From Natural Language Supervision},
  booktitle    = {ICML},
  volume       = {139},
  pages        = {8748--8763},
  year         = {2021}
}

@inproceedings{iccv23/SigLIP,
  author       = {Xiaohua Zhai and
                  Basil Mustafa and
                  Alexander Kolesnikov and
                  Lucas Beyer},
  title        = {Sigmoid Loss for Language Image Pre-Training},
  booktitle    = {ICCV},
  pages        = {11941--11952},
  year         = {2023}
}

@inproceedings{cvpr19/GQA,
  author       = {Drew A. Hudson and
                  Christopher D. Manning},
  title        = {{GQA:} {A} New Dataset for Real-World Visual Reasoning and Compositional Question Answering},
  booktitle    = {CVPR},
  pages        = {6700--6709},
  year         = {2019}
}

@inproceedings{cvpr17/VQAv2,
  author       = {Yash Goyal and
                  Tejas Khot and
                  Douglas Summers{-}Stay and
                  Dhruv Batra and
                  Devi Parikh},
  title        = {Making the {V} in {VQA} Matter: Elevating the Role of Image Understanding in Visual Question Answering},
  booktitle    = {CVPR},
  pages        = {6325--6334},
  year         = {2017}
}

@inproceedings{cvpr19/textvqa,
  author       = {Amanpreet Singh and
                  Vivek Natarajan and
                  Meet Shah and
                  Yu Jiang and
                  Xinlei Chen and
                  Dhruv Batra and
                  Devi Parikh and
                  Marcus Rohrbach},
  title        = {Towards {VQA} Models That Can Read},
  booktitle    = {CVPR},
  pages        = {8317--8326},
  year         = {2019}
}

@inproceedings{emnlp23/POPE,
  author       = {Yifan Li and
                  Yifan Du and
                  Kun Zhou and
                  Jinpeng Wang and
                  Wayne Xin Zhao and
                  Ji{-}Rong Wen},
  title        = {Evaluating Object Hallucination in Large Vision-Language Models},
  booktitle    = {EMNLP},
  pages        = {292--305},
  year         = {2023}
}

@inproceedings{eccv24/mmbench,
  author       = {Yuan Liu and
                  Haodong Duan and
                  Yuanhan Zhang and
                  Bo Li and
                  Songyang Zhang and
                  Wangbo Zhao and
                  Yike Yuan and
                  Jiaqi Wang and
                  Conghui He and
                  Ziwei Liu and
                  Kai Chen and
                  Dahua Lin},
  title        = {MMBench: Is Your Multi-modal Model an All-Around Player?},
  booktitle    = {ECCV},
  volume       = {15064},
  pages        = {216--233},
  year         = {2024}
}

@article{arxiv24/MME,
      title={MME: A Comprehensive Evaluation Benchmark for Multimodal Large Language Models}, 
      author={Chaoyou Fu and Peixian Chen and Yunhang Shen and Yulei Qin and Mengdan Zhang and Xu Lin and others},
      journal={arXiv preprint arXiv:2306.13394},
      year={2024}
}

@inproceedings{cvpr18/Vizwiz,
  author       = {Danna Gurari and
                  Qing Li and
                  Abigale J. Stangl and
                  Anhong Guo and
                  Chi Lin and
                  Kristen Grauman and
                  Jiebo Luo and
                  Jeffrey P. Bigham},
  title        = {VizWiz Grand Challenge: Answering Visual Questions From Blind People},
  booktitle    = {CVPR},
  pages        = {3608--3617},
  year         = {2018}
}

@inproceedings{neurips22/sqa,
  author       = {Pan Lu and
                  Swaroop Mishra and
                  Tanglin Xia and
                  Liang Qiu and
                  Kai{-}Wei Chang and
                  Song{-}Chun Zhu and
                  Oyvind Tafjord and
                  Peter Clark and
                  Ashwin Kalyan},
  title        = {Learn to Explain: Multimodal Reasoning via Thought Chains for Science Question Answering},
  booktitle    = {NeurIPS},
  year         = {2022}
}

@inproceedings{cvpr24/LION,
  author       = {Gongwei Chen and
                  Leyang Shen and
                  Rui Shao and
                  Xiang Deng and
                  Liqiang Nie},
  title        = {{LION} : Empowering Multimodal Large Language Model with Dual-Level Visual Knowledge},
  booktitle    = {CVPR},
  pages        = {26530--26540},
  year         = {2024}
}

@inproceedings{neurips17/attention,
  author       = {Ashish Vaswani and
                  Noam Shazeer and
                  Niki Parmar and
                  Jakob Uszkoreit and
                  Llion Jones and
                  Aidan N. Gomez and
                  Lukasz Kaiser and
                  Illia Polosukhin},
  title        = {Attention is All you Need},
  booktitle    = {NeurIPS},
  pages        = {5998--6008},
  year         = {2017}
}

@article{arxiv24/Gemma,
      title={Gemma: Open Models Based on Gemini Research and Technology}, 
      author={Gemma Team and Thomas Mesnard and Cassidy Hardin and Robert Dadashi and Surya Bhupatiraju and Shreya Pathak and others},
      journal={arXiv preprint arXiv:2403.08295},
      year={2024}
}

@inproceedings{iccv25/VisPruner,
      title={Beyond Text-Visual Attention: Exploiting Visual Cues for Effective Token Pruning in VLMs}, 
      author={Qizhe Zhang and Aosong Cheng and Ming Lu and Renrui Zhang and Zhiyong Zhuo and Jiajun Cao and Shaobo Guo and Qi She and Shanghang Zhang},
      booktitle = {ICCV},
      year={2025}
}

@inproceedings{emnlp24/videollava,
  author       = {Bin Lin and
                  Yang Ye and
                  Bin Zhu and
                  Jiaxi Cui and
                  Munan Ning and
                  Peng Jin and
                  Li Yuan},
  title        = {Video-LLaVA: Learning United Visual Representation by Alignment Before Projection},
  booktitle    = {EMNLP},
  pages        = {5971--5984},
  year         = {2024}
}

@inproceedings{cvpr17/tgif,
  author       = {Yunseok Jang and
                  Yale Song and
                  Youngjae Yu and
                  Youngjin Kim and
                  Gunhee Kim},
  title        = {{TGIF-QA:} Toward Spatio-Temporal Reasoning in Visual Question Answering},
  booktitle    = {CVPR},
  pages        = {1359--1367},
  year         = {2017}
}

@inproceedings{acl11/msvd,
  author       = {David L. Chen and
                  William B. Dolan},
  title        = {Collecting Highly Parallel Data for Paraphrase Evaluation},
  booktitle    = {ACL},
  pages        = {190--200},
  year         = {2011}
}

@inproceedings{cvpr16/msrvtt,
  author       = {Jun Xu and
                  Tao Mei and
                  Ting Yao and
                  Yong Rui},
  title        = {{MSR-VTT:} {A} Large Video Description Dataset for Bridging Video and Language},
  booktitle    = {CVPR},
  pages        = {5288--5296},
  year         = {2016}
}

@inproceedings{eccv14/mscoco,
  author       = {Tsung{-}Yi Lin and
                  Michael Maire and
                  Serge J. Belongie and
                  James Hays and
                  Pietro Perona and
                  Deva Ramanan and
                  Piotr Doll{\'{a}}r and
                  C. Lawrence Zitnick},
  title        = {Microsoft {COCO:} Common Objects in Context},
  booktitle    = {ECCV},
  volume       = {8693},
  pages        = {740--755},
  year         = {2014}
}

@article{openimages,
  title={OpenImages: A public dataset for large-scale multi-label and multi-class image classification.},
  author={Krasin, Ivan and Duerig, Tom and Alldrin, Neil and Ferrari, Vittorio and Abu-El-Haija, Sami and Kuznetsova, Alina and others},
  journal={Dataset available from https://github.com/openimages},
  year={2017}
}

@article{ijcv17/visualgenome,
  author       = {Ranjay Krishna and
                  Yuke Zhu and
                  Oliver Groth and
                  Justin Johnson and
                  Kenji Hata and
                  Joshua Kravitz and
                  others},
  title        = {Visual Genome: Connecting Language and Vision Using Crowdsourced Dense Image Annotations},
  journal      = {Int. J. Comput. Vis.},
  volume       = {123},
  number       = {1},
  pages        = {32--73},
  year         = {2017}
}

@article{arxiv25/VmapL,
      title={How Visual Representations Map to Language Feature Space in Multimodal LLMs}, 
      author={Constantin Venhoff and Ashkan Khakzar and Sonia Joseph and Philip Torr and Neel Nanda},
      year={2025},
      journal={arXiv preprint arXiv:2506.11976}
}

@inproceedings{
neurips25/vla-cache,
title={{VLA}-Cache: Efficient Vision-Language-Action Manipulation via Adaptive Token Caching},
author={Siyu Xu and Yunke Wang and Chenghao Xia and Dihao Zhu and Tao Huang and Chang Xu},
booktitle={NeurIPS},
year={2025}
}

\newpage
\begin{center}
    \LARGE \textbf{Supplementary Material} \par
\end{center}

\appendix

\section{Methodology}

Algorithm~\ref{algo:visiondrop} details the token reduction procedure of VisionDrop at a single stage, including dominant token selection and contextual token merging.

\begin{algorithm}[h]
\caption{Pseudocode at Single Stage}
\label{algo:visiondrop}
\algcomment{\fontsize{9pt}{0em}\selectfont \texttt{uni\_split}: Uniformly sample references, others used for merging; \texttt{avg\_merge}: Merge tokens into their assigned references.
}
\lstset{
  backgroundcolor=\color{white},
  basicstyle=\fontsize{9pt}{9pt}\ttfamily\selectfont,
  columns=fullflexible,
  breaklines=true,
  captionpos=b,
  commentstyle=\fontsize{9pt}{9pt},
  keywordstyle=\fontsize{9pt}{9pt},
  numbers=none,
  xleftmargin=0pt,       
  framexleftmargin=0pt,  
}
\begin{lstlisting}[language=python]
# Q: visual query states; q: query length
# K: key states
# X: visual tokens; v: visual token length
# s: sequence length; d: hidden dimension
# h: number of attention heads
# n_dom: number of dominant tokens
# n_ctx: number of contextual tokens
# IDX: index of visual tokens

# attn in shape (h, q, s)
attn = softmax(Q @ K.T / sqrt(d))
# Average over heads
attn = attn.mean(0) # (q, s)
# Select visual part
attn = att[:, IDX] # (q, v)
# Average over queries
attn_scores = attn.mean(0) # (v, )

# Dominant Token Selection
topk_idx = attn_scores.topk(n_dom).indices
dominant_tokens = X[topk_idx]

# Select remaining tokens
remaining = X[~topk_idx] # (v-n_dom, )

# Split into reference and candidate sets
ref, cand = uni_split(remaining, n_ctx)

K = K.mean(0)[IDX, :] # (v, d)
K = normalize(K[~topk_idx]) # (v-n_dom, )

# Split metrics
K_ref, K_cand = uni_split(K, n_ctx)

# Compute similarity based on the metrics
similarity = K_cand @ K_ref.T

# Assign each candidate to the most similar reference
assign_idx = similarity.argmax(dim=1)

# Contextual Token Merging
contextual_tokens = avg_merge(assign_idx, ref, cand)

X_out = concat(dominant_tokens, contextual_tokens)
\end{lstlisting}
\end{algorithm}


\section{Experimental Setups}

\subsection{Benchmarks and Metrics}
We evaluate our method on 12 widely used benchmarks, including 9 image understanding benchmarks and 3 video understanding benchmarks. All evaluations follow the original protocols and settings used in LLaVA-series~\cite{cvpr24/LLaVA1.5, liu2024llavanext, emnlp24/videollava}.

\subsubsection{Image Benchmarks.} 
We adopt 9 image benchmarks, all formulated as vision question answering (VQA) tasks, with varying levels of multimodal reasoning complexity.

\textit{GQA.} The GQA~\cite{cvpr19/GQA} benchmark evaluates the model’s ability to perform compositional reasoning and structured scene understanding. Each image is paired with multiple question-answer pairs generated from a corresponding scene graph~\cite{ijcv17/visualgenome}, which encodes objects, attributes, and their spatial or semantic relationships. The questions are synthesized using a functional program over the scene graph. We follow the standard evaluation protocol and report accuracy on the test-dev split, which includes 12,578 image-question pairs.

\textit{MMBench.} The MMBench~\cite{eccv24/mmbench} benchmark evaluates the general capabilities of multimodal models through a hierarchical skill taxonomy. The evaluation consists of three levels: the top level focuses on fundamental perceptual and reasoning abilities, the middle level divides these into six skill categories, and the lowest level includes 20 fine-grained task types that capture diverse multimodal challenges. Each task uses carefully designed multiple-choice questions to target specific skills. We use the development sets provided by OpenCompass, each containing 6,666 image-question pairs in English or Chinese (referred to as MMBench-CN). Model performance is measured using accuracy.

\textit{MME.} The MME~\cite{arxiv24/MME} benchmark evaluates multimodal models on visual perception and cognitive reasoning across 14 distinct subtasks. These tasks cover a range from optical character recognition (OCR) to both coarse and fine-grained visual understanding. Coarse-level tasks assess object count, spatial relationships, and basic attributes like color, while fine-grained tasks involve recognizing specific entities such as celebrities, landmarks, scenes, posters, and artworks. All tasks are formulated as binary classification problems. The overall performance is measured using the MME score, which aggregates accuracy across all subtasks. The benchmark includes 2,374 image-question pairs.

\textit{POPE.} The POPE~\cite{emnlp23/POPE} benchmark focuses on evaluating object hallucination in vision-language models. It formulates the task as a set of binary questions about whether specific objects appear in images, providing a direct measure of hallucination. Images are sourced from the MSCOCO~\cite{eccv14/mscoco} dataset, and evaluation is based on average accuracy across three sampling strategies, covering 8,910 image-question pairs.

\textit{ScienceQA.} ScienceQA~\cite{neurips22/sqa} is a benchmark that evaluates zero-shot performance on scientific multiple-choice questions. It covers 3 subject areas, including natural science, social science, and language science. The questions are organized into 26 topics, 127 categories, and 379 skills. Many questions are paired with supporting illustrations, though not all include images. For our evaluation, we use the SQA-IMG subset of the test set, which contains 2,017 questions each accompanied by an image. Model performance is measured using accuracy.

\textit{VQAv2.} VQAv2~\cite{cvpr17/VQAv2} is designed to evaluate a model’s ability to answer open-ended questions based on visual input. It includes 265,016 images from the MSCOCO~\cite{eccv14/mscoco} dataset, with each image paired with at least three human-annotated questions. The dataset introduces an adversarial balancing strategy, where each question is linked to multiple images with different correct answers, reducing the model’s reliance on dataset biases. For evaluation, we use the test-dev split, which contains 107,394 image-question pairs. Accuracy is computed based on agreement with 10 human-labeled answers.

\textit{TextVQA.} The TextVQA~\cite{cvpr19/textvqa} benchmark evaluates a model’s ability to recognize and reason over textual information in images. It targets real-world contexts where textual information is visually presented, such as in signs, advertisements, and packaging. The images are mainly sourced from the Open Images v3~\cite{openimages} dataset, and are accompanied by OCR token annotations to facilitate text recognition. Answering questions may involve direct reading or require integrating textual cues with visual context. We evaluate model performance using accuracy on the validation set, which includes 5,000 image-question pairs.

\textit{VizWiz.} VizWiz~\cite{cvpr18/Vizwiz} is a VQA benchmark grounded in real-world scenarios, featuring images captured by blind individuals. Each image is paired with a naturally posed question from the user, and corresponding answers are gathered from crowdworkers, yielding 10 human annotations per question. The dataset presents practical challenges such as image blur, poor lighting, and occasionally mismatched or underspecified queries. We evaluate model performance on the test-dev split, which contains 8,000 image-question pairs, using accuracy as the evaluation metric.

\subsubsection{Video Benchmarks.}
To test performance under higher visual redundancy, we include three video-based VQA benchmarks.

\textit{TGIF.} The TGIF~\cite{cvpr17/tgif} benchmark extends VQA to the video domain by introducing tasks that require models to reason over temporal patterns and dynamic scenes. It includes challenges such as repetition counting, action recognition, and state transitions, as well as visual questions grounded in short animated clips.

\textit{MSVD.} The MSVD~\cite{acl11/msvd} benchmark consists of short, diverse video clips paired with natural language questions derived from descriptive captions. It evaluates the ability of models to generate open-ended answers grounded in everyday visual scenarios and language understanding.

\textit{MSRVTT.} The MSRVTT~\cite{cvpr16/msrvtt} benchmark features a large and varied collection of video clips accompanied by open-ended questions. It is designed to test comprehensive video understanding, with a focus on both visual perception and temporal reasoning across a broad spectrum of real-world content.

Following previous works~\cite{emnlp24/videollava,eccv24/FastV}, we use GPT-3.5-Turbo to evaluate model responses on the first 1,000 samples of each benchmark. Each response is scored based on accuracy, a binary judgment indicating correctness, and quality score, an integer from 0 to 5 reflecting the response's relevance and informativeness.

\begin{figure*}[!t]
  \centering
  \includegraphics[width=0.76\textwidth]{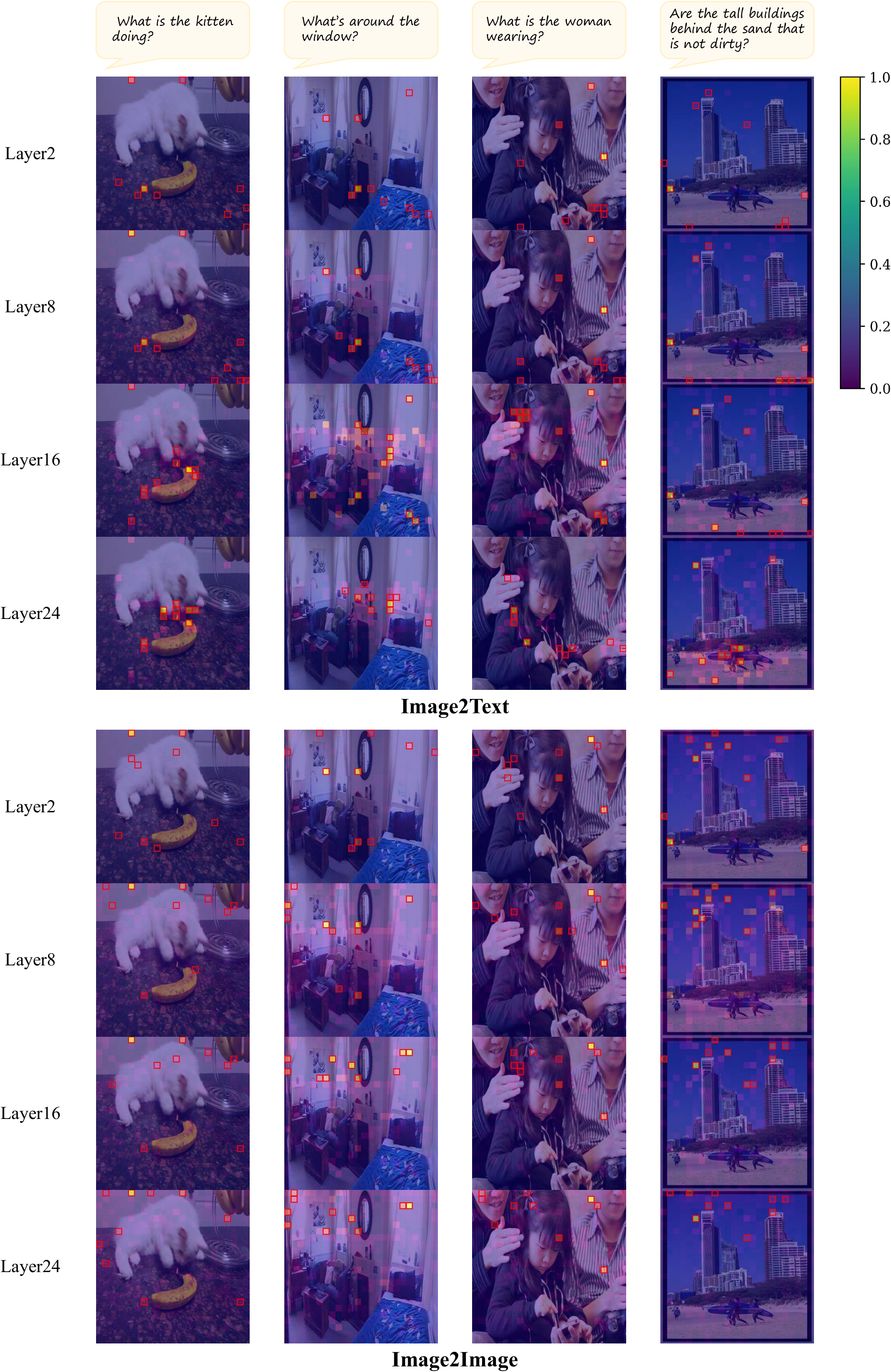}
  \caption{Visualization of image-to-text (top 4 rows) and image-to-image (bottom 4 rows) attention maps at the 2nd, 8th, 16th, and 24th LLM layers in LLaVA-1.5-7B~\cite{cvpr24/LLaVA1.5}. The colorbar indicates normalized attention weights, ranging from 0.0 (low) to 1.0 (high). Red boxes highlight the top-10 attended patches. Notably, image-to-image attention consistently attends to a stable set of informative tokens across layers.}
  \label{fig:ab_attention}
\end{figure*}

\begin{table*}[!t]
\centering
\small
\setlength{\tabcolsep}{9.5pt}
\begin{tabular}{l|cccccccccc}
\toprule
\textbf{Method}         & \textbf{GQA} & \textbf{MMB} & \textbf{MMB}$^{\text{CN}}$ & \textbf{MME} & \textbf{POPE} & \textbf{SQA} & \textbf{VQA}$^{\text{v2}}$ & \textbf{VQA}$^{\text{Text}}$ & \textbf{VizWiz} & \textbf{Avg.} \\
\hline
\rowcolor{gray!15}
\multicolumn{11}{c}{\textit{Upper Bound, 576 Tokens (100\%)}} \\
LLaVA-1.5               & 61.92 & 66.31 & 58.63 & 1863 & 86.81 & 69.51 & 78.53 & 58.20 & 50.13 & 100\% \\
\hline
\rowcolor{gray!15}
\multicolumn{11}{c}{\textit{Retain 192 Tokens (\textdownarrow 66.7\%)}} \\
Causal                  & 59.99 & 65.19 & 58.41 & 1801 & 87.23 & 69.06 & 77.28 & 57.81 & 50.01 & 98.76\% \\
Non-causal              & 59.79 & 64.97 & 57.79 & 1767 & 87.05 & 68.57 & 77.20 & 57.77 & 50.24 & 98.30\% \\
\hline
\rowcolor{gray!15}
\multicolumn{11}{c}{\textit{Retain 128 Tokens (\textdownarrow 77.8\%)}} \\
Causal                  & 58.61 & 64.52 & 57.06 & 1777 & 85.92 & 68.52 & 76.24 & 57.63 & 51.06 & 97.80\% \\
Non-causal              & 58.95 & 64.13 & 57.62 & 1780 & 86.34 & 68.32 & 76.41 & 57.43 & 51.27 & 97.97\% \\
\hline
\rowcolor{gray!15}
\multicolumn{11}{c}{\textit{Retain 64 Tokens (\textdownarrow 88.9\%)}} \\
Causal                  & 55.89 & 62.95 & 55.10 & 1698 & 81.58 & 69.31 & 73.16 & 55.59 & 52.28 & 95.22\% \\
Non-causal              & 56.10 & 62.72 & 55.55 & 1704 & 82.07 & 68.77 & 73.71 & 56.07 & 52.94 & 95.63\% \\
\hline
\rowcolor{gray!15}
\multicolumn{11}{c}{\textit{Retain 32 Tokens (\textdownarrow 94.4\%)}} \\
Causal                  & 52.79 & 60.31 & 52.91 & 1572 & 77.19 & 69.41 & 68.55 & 53.56 & 52.26 & 91.46\% \\
Non-causal              & 53.04 & 59.98 & 52.30 & 1572 & 77.71 & 69.11 & 69.00 & 53.83 & 52.33 & 91.49\% \\
\bottomrule
\end{tabular}
\caption{Ablation of visual causal attention on LLaVA-1.5-7B~\cite{cvpr24/LLaVA1.5} under different token retention rates.}
\label{tab:ab_attention}
\end{table*}

\subsection{Implementation Details}
All experiments are conducted on the NVIDIA RTX 3090 GPU with 24GB memory under Ubuntu 20.04. Our implementation uses Python 3.10 and builds on PyTorch and HuggingFace Transformers. All required dependencies and runtime settings are listed in the pyproject.toml, which are included in our released code for full reproducibility.

\section{More Discussions}

\subsection{Analysis of Attention Distribution}
We visualize and compare the image-to-text (top 4 rows) and image-to-image (bottom 4 rows) attention maps at the 2nd, 8th, 16th, and 24th layers of the LLM in LLaVA-1.5-7B~\cite{cvpr24/LLaVA1.5} in Figure~\ref{fig:ab_attention}. In standard visual encoders such as CLIP~\cite{icml21/CLIP} and SigLIP~\cite{iccv23/SigLIP}, it has been observed that—aside from special tokens like \texttt{[CLS]}—a small subset of visual tokens tends to receive significantly higher attention and encode most of the information in deeper layers, with the majority receiving minimal attention~\cite{cvpr25/VisionZip}. Notably, in Figure~\ref{fig:ab_attention}, image-to-image attention consistently preserves a small set of informative visual tokens across layers. This behavior aligns with the redundancy observed in visual encoders, indicating that visual information is also aggregated in LLM layers. In contrast, image-to-text attention, despite at times focusing on text-relevant regions, suffers from the misalignment issues discussed in the main paper and fails to retain such dominant tokens, resulting in information loss. Furthermore, as the LLM depth increases, attention becomes increasingly entangled in both attention types, underscoring the importance of preserving the informative tokens throughout the LLM layers.

\subsection{Analysis of Visual Attention}
In LLaVA-series~\cite{cvpr24/LLaVA1.5, liu2024llavanext, emnlp24/videollava}, causal attention is uniformly applied to both visual and textual tokens, despite the inherently bidirectional nature of visual information. Since our method is training-free and built on pretrained models, we retain the causal mask when computing visual-only attention to stay consistent with the original model. To further investigate whether this constraint affects visual token scoring, we compare the default causal setting with a variant that removes the causal mask in visual-to-visual attention. As shown in Table~\ref{tab:ab_attention}, the non-causal variant achieves comparable or slightly superior performance across all token retention rates, even though the upper triangular region of the visual attention matrix is untrained. These findings underscore the distinct characteristics of visual modalities in contrast to language. The fact that causal masking can be safely removed without performance degradation further questions the appropriateness of text-guided visual token reduction within LLM, and supports the use of our vision-only scoring strategy.

\end{document}